\documentclass[10pt,twocolumn,letterpaper]{article}

\usepackage{cvpr}
\usepackage{times}
\usepackage{epsfig}
\usepackage{graphicx}
\usepackage{amsmath}
\usepackage{amssymb}
\usepackage{listings}
\usepackage{url}

\usepackage[breaklinks=true,bookmarks=false]{hyperref}

\cvprfinalcopy %

\def\cvprPaperID{****} %

\begin{document}

\title{HoloLens 2 Research Mode as a Tool for Computer Vision Research}

\author{Dorin Ungureanu \hspace{0.03\linewidth} Federica Bogo \hspace{0.03\linewidth} Silvano Galliani \hspace{0.03\linewidth} Pooja Sama \hspace{0.03\linewidth} Xin Duan \hspace{0.03\linewidth}\\
Casey Meekhof \hspace{0.03\linewidth} Jan St\"{u}hmer\thanks{Now at Samsung AI Centre, Cambridge (UK). Work performed while at Microsoft.} \hspace{0.03\linewidth} Thomas J. Cashman \hspace{0.03\linewidth} Bugra Tekin\\
Johannes L. Sch\"{o}nberger \hspace{0.03\linewidth} Pawel Olszta \hspace{0.03\linewidth} Marc Pollefeys\\
Microsoft
}

\maketitle

\begin{abstract}
Mixed reality headsets, such as the Microsoft HoloLens~2, are powerful sensing devices with integrated compute capabilities, which makes it an ideal platform for computer vision research.
In this technical report, we present HoloLens~2 Research Mode, an API and a set of tools enabling access to the raw sensor streams.
We provide an overview of the API and explain how it can be used to build mixed reality applications based on processing sensor data.
We also show how to combine the Research Mode sensor data with the built-in eye and hand tracking capabilities provided by HoloLens~2.
By releasing the Research Mode API and a set of open-source tools, we aim to foster further research in the fields of computer vision as well as robotics and encourage contributions from the research community.
\end{abstract}

\section{Introduction}
\label{ref:intro}
Mixed reality technologies have tremendous potential to fundamentally transform the way we interact with our environment, with other people, and the physical world.
Mixed reality headsets such as the Microsoft HoloLens 1 \& 2 have already seen great adoption in a number of fields, particularly in first-line worker scenarios, ranging from assisted surgery to remote collaboration and from task guidance to overlaying digital twin on the real world. 
Despite existing adoption in these fields, the general space of mixed reality is still in its infancy.
Oftentimes, the development of new mixed reality applications requires fundamental research and the novel combination of different sensors.
The entry barrier to computer vision research in this area is significantly lowered by access to tools enabling us to effectively collect raw sensor data and develop new computer vision algorithms that run on-device.

The first-generation HoloLens Research Mode, released in 2018, enabled computer vision research on device by providing access to all raw image sensor streams -- including depth and IR.
Released together with a public repository collecting auxiliary tools and sample applications~\cite{hlforcv}, Research Mode promoted the use of HoloLens as a powerful tool for doing research in computer vision and robotics~\cite{workshop}.

\begin{figure}[t]
   \begin{center}
      \includegraphics[width=1.\linewidth]{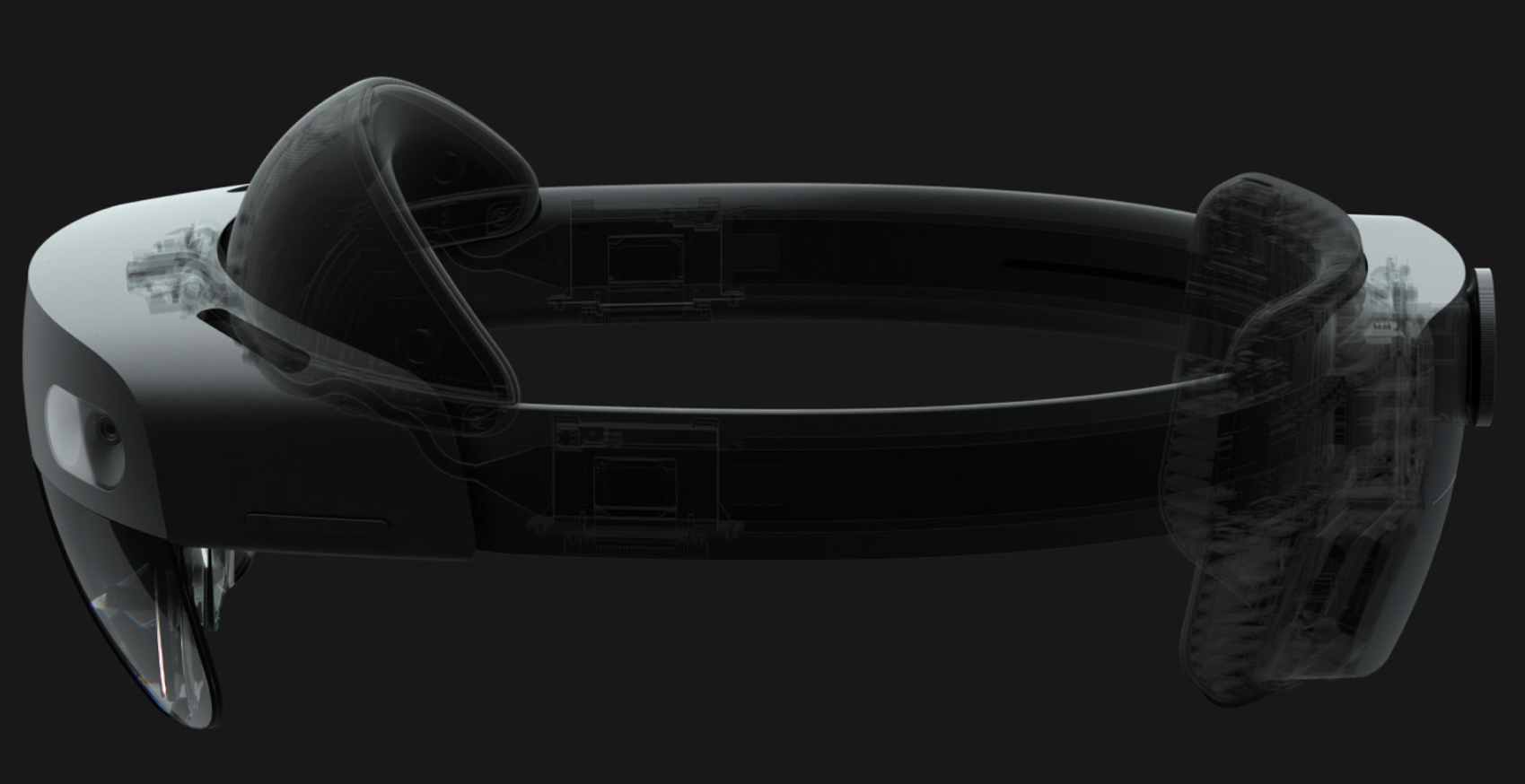}
   \end{center}
      \caption{Microsoft HoloLens~2.}
   \label{fig:hl2}
   \end{figure}

HoloLens 2 (Fig.~\ref{fig:hl2}), announced in 2019, brings a number of improvements with respect to the first-generation device -- like a dedicated DNN core, articulated hand tracking and eye gaze tracking~\cite{hl2}.
However, being a novel platform built on new hardware, HoloLens~2 is not compatible with the previous version of Research Mode.

In this technical report, we introduce the second-generation HoloLens Research Mode, made available in 2020.
Research Mode provides a set of C++ APIs and tools to access the HoloLens~2 sensor streams.
We discuss the main novelties with respect to the previous version and present a set of applications built on top of it.
For additional material and API documentation, we refer the reader to the GitHub repository~\cite{hl2forcv}.

Research Mode is designed for academic and industrial researchers exploring new ideas in the fields of computer vision and robotics. It is not intended for applications deployed to end-users.
Additionally, Microsoft does not provide assurances that Research Mode will be supported in future hardware or OS updates.

The rest of this technical report is organized as follows.
Section~\ref{sec:hl2} introduces the HoloLens~2 device, detailing in particular its input streams.
Section~\ref{sec:api} provides an overview of the Research Mode API, while Section~\ref{sec:examples} showcases a few example applications.
Finally, Section~\ref{sec:conclusion} sums up our contributions.

\section{HoloLens 2}
\label{sec:hl2}
The HoloLens~2 mixed reality headset brings a set of improvements with respect to the first-generation device -- including a larger field of view, a custom DNN core, fully articulated hand tracking and eye gaze tracking.

\begin{figure}[t]
   \begin{center}
      \includegraphics[width=1.\linewidth]{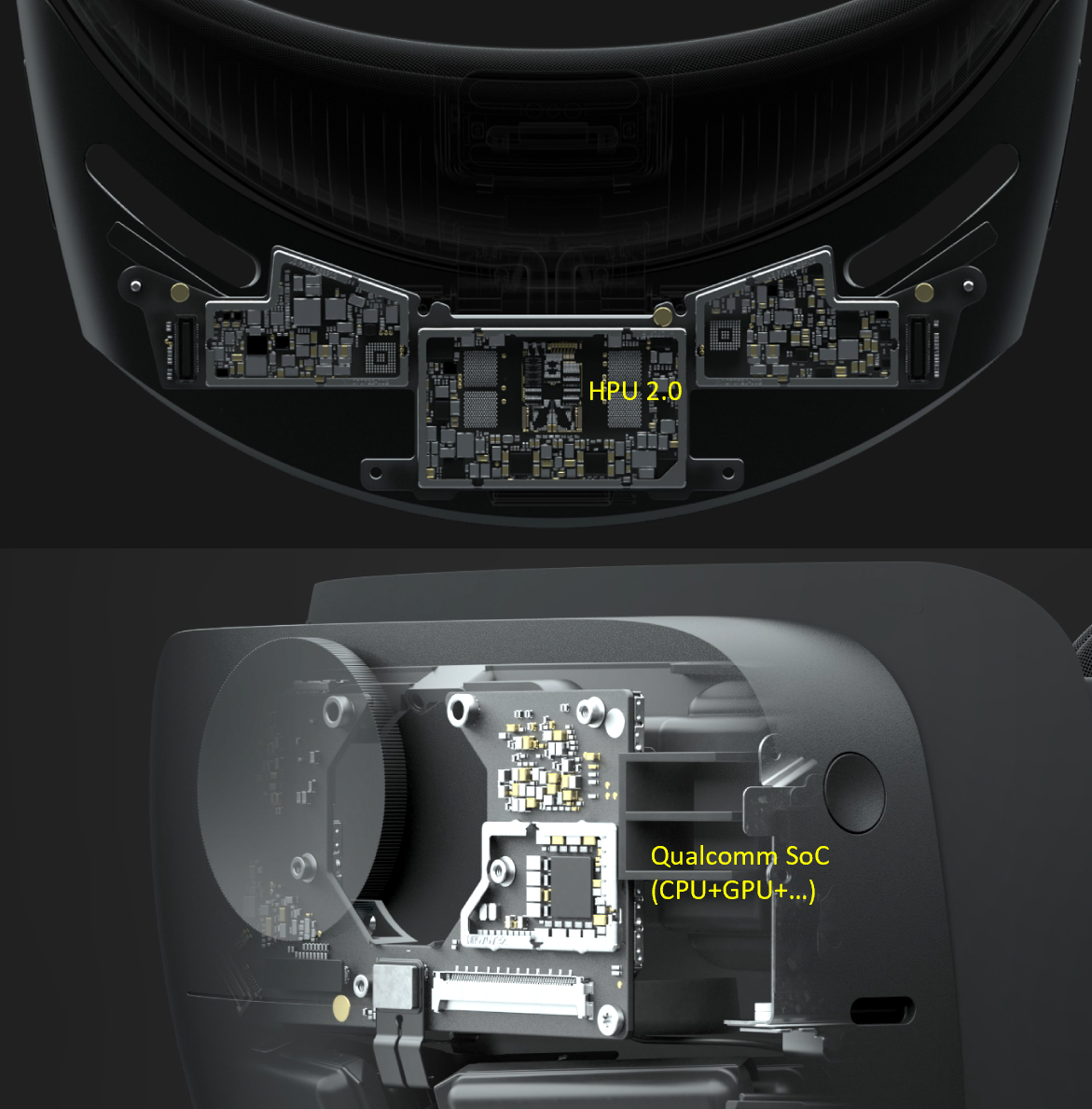}
   \end{center}
      \caption{HPU and SoC: The HPU (top) is located on the frontal part of the device, near the sensors.
      The SoC (bottom), a Qualcomm Snapdragon 850, is located on the rear.}
   \label{fig:hpu}
   \end{figure}

The device features a second-generation custom-built Holographic Processing Unit (HPU 2.0), which enables low-power, real-time computer vision.
The HPU runs all the computer vision algorithms on device (head tracking, hand tracking, eye gaze tracking, spatial mapping etc.) and hosts the DNN core.
It is located on the front part of the device, near the sensors (Fig.~\ref{fig:hpu}, top).
The CPU on the SoC (a Qualcomm SnapDragon 850) remains fully available for applications. The SoC is located on the rear (Fig.~\ref{fig:hpu}, bottom).

\begin{figure}[t]
   \begin{center}
      \includegraphics[width=1.\linewidth]{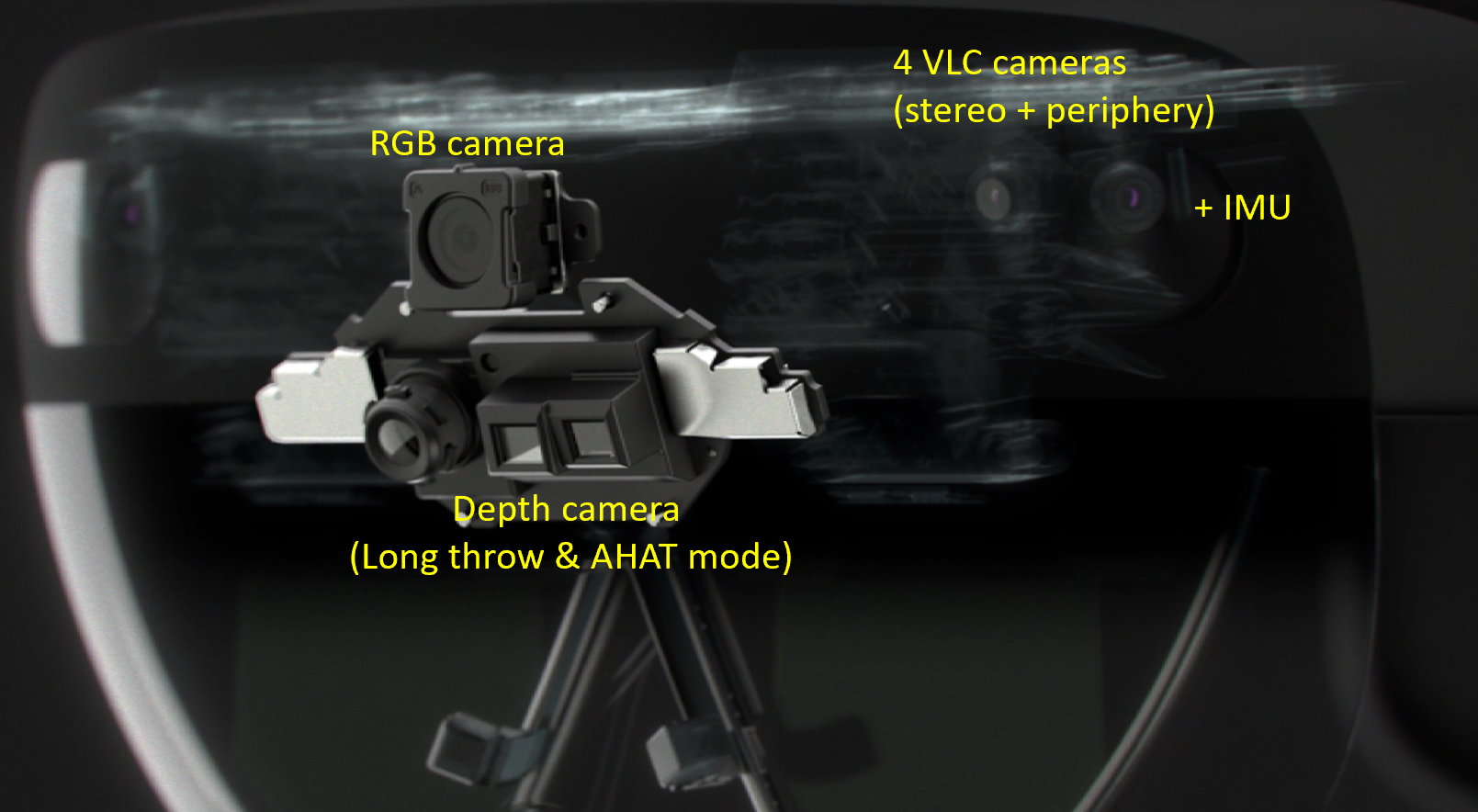}
   \end{center}
      \caption{HoloLens~2 input sensors.}
   \label{fig:cameras}
   \end{figure}

The device is equipped with a depth and an RGB camera, four grayscale cameras, and an Inertial Measurement Unit (IMU), as shown in Fig.~\ref{fig:cameras}.
Audio is captured with a microphone array (5 channels).

\begin{figure}[t]
   \begin{center}
      \includegraphics[width=1.\linewidth]{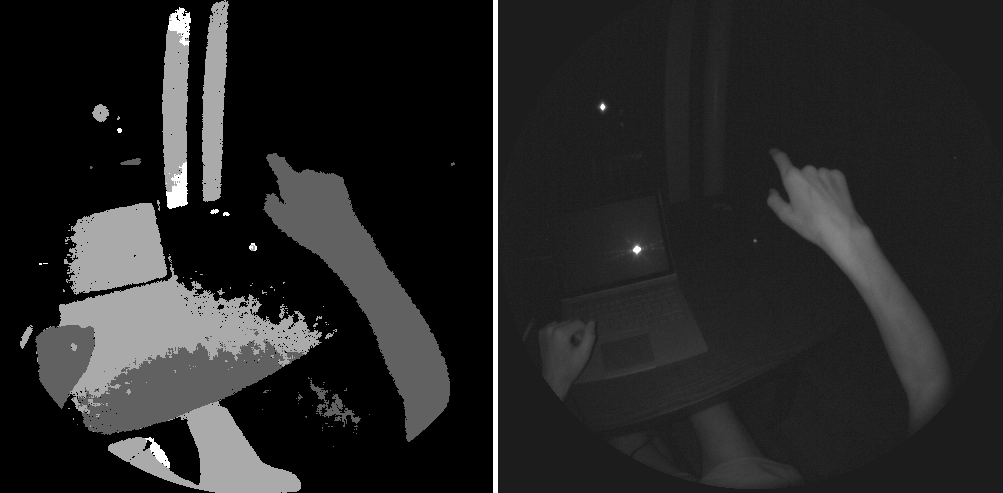}
   \end{center}
      \caption{Depth in AHAT mode: Depth (left) and Active Brightness (right) images.}
   \label{fig:ahat}
   \end{figure}

Research Mode for HoloLens~2 enables access to the following input streams:
\begin{itemize}
\item Four visible-light tracking cameras (VLC): Grayscale cameras (30 fps) used by the system for real-time visual-inertial SLAM. 
\item A depth camera, which operates in two modes:
   \begin{itemize}
      \item AHAT (Articulated HAnd Tracking), high-framerate (45 fps) near-depth sensing used for hand tracking. As hands are supported up to 1 meter from the device, the HoloLens 2 saves power by calculating only ``aliased depth'' from the phase-based time of flight camera.
            This means that the signal contains only the fractional part of the distance from the device when expressed in meters (see Fig.~\ref{fig:ahat}).
      \item Long Throw, low-framerate (1-5 fps) far-depth sensing used to compute spatial mapping on device. 
   \end{itemize}
\item Two depth modes of the IR stream (Active Brightness, AB in short), computed from the same modulated IR signal for depth computation. These images are illuminated by infrared and unaffected by ambient visible light (see Fig.~\ref{fig:ahat}, right).
\item Inertial Measurement Unit (IMU):
\begin{itemize}
   \item Accelerometer, used by the system to determine the linear acceleration along the $x$, $y$ and $z$ axes as well as gravity.
   \item Gyroscope, used by the system to determine rotations.
   \item Magnetometer, used by the system for absolute orientation estimation.
\end{itemize}   
\end{itemize}
For each stream, Research Mode provides interfaces to retrieve frames and associated information (\eg, resolution and timestamps), and to map the sensors with respect to the device and to the world.
Table~\ref{tab:frames} summarizes the main characteristics of each (camera) input stream.
In the following section, we provide an overview of the API.

\begin{table}[t]
   \centering
   \begin{tabular}{|c c c|} 
    \hline
    Stream & Resolution & Format \\ [0.5ex] 
    \hline
    VLC & $640 \times 480$ & 8-bit  \\ 
    Long throw depth & $320 \times 288$ & 16-bit \\
    Long throw AB & $320 \times 288$ & 16-bit \\
    AHAT & $512 \times 512$ & 16-bit \\
    AHAT AB & $512 \times 512$ & 16-bit \\ [1ex] 
    \hline
   \end{tabular}
   \caption{Research Mode camera frames.}
   \label{tab:frames}
\end{table}

\section{Research Mode API}
\label{sec:api}
In this section, we describe the main objects exposed by Research Mode and how they are used to access the sensor input streams. We refer the reader to~\cite{hl2forcv} for a detailed API documentation.

The Research Mode main sensor loop (Sec.~\ref{sec:loop}) starts by creating a {\small\ttfamily ResearchModeDevice} object,
which is used to obtain a list of available {\small\ttfamily Sensors}.
{\small\ttfamily Sensor} objects expose methods to retrieve and process frames (Sec.~\ref{sec:sensors})
and to locate the sensors with respect to the device and the world (Sec.~\ref{sec:coord}).

\subsection{Main Sensor Loop}
\label{sec:loop}
The main sensor processing loop involves instantiating a {\small\ttfamily ResearchModeDevice}, getting sensor descriptors,
opening sensor streams and fetching frames:
\begin{lstlisting}
HRESULT hr = S_OK;
IResearchModeSensorDevice *pSensorDevice;
std::vector<ResearchModeSensorDescriptor> sensorDescriptors;
size_t sensorCount = 0;

hr = CreateResearchModeSensorDevice(&pSensorDevice);

pSensorDevice->DisableEyeSelection();

hr = pSensorDevice->GetSensorCount(&sensorCount);
sensorDescriptors.resize(sensorCount);
hr = pSensorDevice->GetSensorDescriptors(
                        sensorDescriptors.data(), 
                        sensorDescriptors.size(),
                        &sensorCount);

for (auto& sensorDescriptor : sensorDescriptors)
{
   // Sensor frame read thread
   IResearchModeSensor *pSensor = nullptr;
   size_t sampleBufferSize;
   IResearchModeSensorFrame* pSensorFrame = nullptr;

   hr = pSensorDevice->GetSensor(
                           sensorDescriptor.sensorType,
                           &pSensor);   
   hr = pSensor->GetSampleBufferSize(&sampleBufferSize);
   hr = pSensor->OpenStream();

   for (UINT i = 0; i < 4; i++)
   {
      hr = pSensor->GetNextBuffer(&pSensorFrame);
      if (pSensor->GetSensorType() >= IMU_ACCEL)
      {
         // Process IMU frame
         SaveFrameImu(pSensor, pSensorFrame, i);
      }
      else
      {
         // Process camera frame
         SaveFrameCamera(pSensor, pSensorFrame, i);
      }
      if (pSensorFrame)      
         pSensorFrame->Release();   
   }
   hr = pSensor->CloseStream();
   if (pSensor)   
      pSensor->Release();   
}
pSensorDevice->EnableEyeSelection();
pSensorDevice->Release();
return hr;
\end{lstlisting}

Note that {\small\ttfamily OpenStream} and {\small\ttfamily GetNextBuffer} need to be called from the same thread.
Since {\small\ttfamily GetNextBuffer} calls are blocking, per-sensor frame loops should be run on their own thread.
This allows sensors to be processed at their own framerate.

\subsection{Sensors and Sensor Frames}
\label{sec:sensors}
The {\small\ttfamily IResearchModeSensor} interface introduced in the previous section abstracts the Research Mode sensors.
It provides methods and properties common to all sensors: {\small\ttfamily OpenStream}, {\small\ttfamily CloseStream}, {\small\ttfamily GetFriendlyName},
{\small\ttfamily GetSensorType}, {\small\ttfamily GetNextBuffer}.
Sensors can be of the following types:
\begin{lstlisting}
enum ResearchModeSensorType
{
      LEFT_FRONT,
      LEFT_LEFT,
      RIGHT_FRONT,
      RIGHT_RIGHT,
      DEPTH_AHAT,
      DEPTH_LONG_THROW,
      IMU_ACCEL,
      IMU_GYRO,
      IMU_MAG
};   
\end{lstlisting}
where the first six types are \emph{camera} sensors, and the remaining three are \emph{IMU} sensors. 
Camera and IMU sensors differ in the methods they expose: for instance, camera sensors expose methods for projecting 3D points in camera space to 2D points in image space (see Sec.~\ref{sec:coord}), while IMU sensors do not.
Sensor specializations are obtained by calling {\small\ttfamily QueryInterface}:
\begin{lstlisting}
IResearchModeSensor* pSensor;
// ... initialize pSensor
IResearchModeCameraSensor* pCameraSensor = nullptr;
HRESULT hr = pSensor->QueryInterface(
                        IID_PPV_ARGS(&pCameraSensor));
\end{lstlisting}

An analogous distinction exists between the {\small\ttfamily IResearchModeSensorFrame} interface and per-sensor specializations.
Once a sensor is in streaming mode, sensor frames are retrieved with {\small\ttfamily IResearchModeSensor::GetNextBuffer}.
All sensor frames have a common {\small\ttfamily IResearchModeSensorFrame} interface that returns frame information common to all types of frames: timestamps and sample size in bytes.
As above, camera and IMU frame specializations are obtained by calling {\small\ttfamily QueryInterface}:
\begin{lstlisting}
IResearchModeSensorFrame* pSensorFrame;
// ... Obtain pSensorFrame via GetNextBuffer

IResearchModeSensorVLCFrame* pVLCFrame = nullptr;
HRESULT hr = pSensorFrame->QueryInterface(
                             IID_PPV_ARGS(&pVLCFrame));
\end{lstlisting}

Camera and IMU frames can be used to access the following information: 
\paragraph{Camera frames} provide getters for resolution, exposure, gain. In addition,
\begin{itemize}
\item VLC frames return grayscale buffers;
\item Long Throw depth frames return a depth buffer, a sigma buffer and an active brightness buffer;
\item AHAT depth frames return a depth buffer and an active brightness buffer.
\end{itemize}
The active brightness buffer returns a so-called IR reading. The value of pixels in the clean IR reading is proportional to the amount of light returned from the scene. The image looks similar to a regular IR image.

The sigma buffer for Long Throw is used to invalidate unreliable depth based on the invalidation mask computed by the depth algorithm.
For AHAT, for efficiency purposes, the invalidation code is embedded in the depth channel itself.

The following code exemplifies how to access Long Throw depth:
\begin{lstlisting}
void ProcessFrame(IResearchModeSensor *pSensor,
                     IResearchModeSensorFrame* pSensorFrame,
                     int bufferCount)
{
   ResearchModeSensorResolution resolution;
   ResearchModeSensorTimestamp timestamp;
   IResearchModeSensorDepthFrame *pDepthFrame = nullptr;
   UINT16 *pAbImage = nullptr;
   UINT16 *pDepth = nullptr;   
   BYTE *pSigma = nullptr;
   // Invalidation mask for Long Throw
   const USHORT mask = 0x80;

   HRESULT hr = S_OK;
   size_t outBufferCount;

   pSensorFrame->GetResolution(&resolution);
   pSensorFrame->GetTimeStamp(&timestamp);

   hr = pSensorFrame->QueryInterface(
                        IID_PPV_ARGS(&pDepthFrame));

   if (SUCCEEDED(hr))
   {
      hr = pDepthFrame->GetSigmaBuffer(&pSigma, &outBufferCount);
      // Process the buffer...
   }
   if (SUCCEEDED(hr))
   {
      // Extract depth buffer
      hr = pDepthFrame->GetBuffer(&pDepth, &outBufferCount);
      // Validate depth
      for (size_t i = 0; i < outBufferCount; ++i)
      {
            // the most significant bit of pSigma[i]
            // tells if the pixel is valid
            bool isInvalid = (pSigma[i] & mask) > 0;
            if (isInvalid)
            {    
               pDepth[i] = 0;
            }
      }
      // Process the buffer...
   }
   if (SUCCEEDED(hr))
   {
      // Extract active brightness buffer
      hr = pDepthFrame->GetAbDepthBuffer(&pAbImage, &outBufferCount);        
      // Process the buffer...
   }
   if (pDepthFrame)
      pDepthFrame->Release();
}  
\end{lstlisting}

\paragraph{IMU frames} store batches of sensor samples. Accelerometer frames store accelerometer measurements and temperature;
gyroscope frames store gyroscope measurements and temperature; magnetometer frames store magnetometer measurements.

\subsection{Sensor Coordinate Frames}
\label{sec:coord}

\begin{figure}[t]
   \begin{center}
      \includegraphics[width=1.\linewidth]{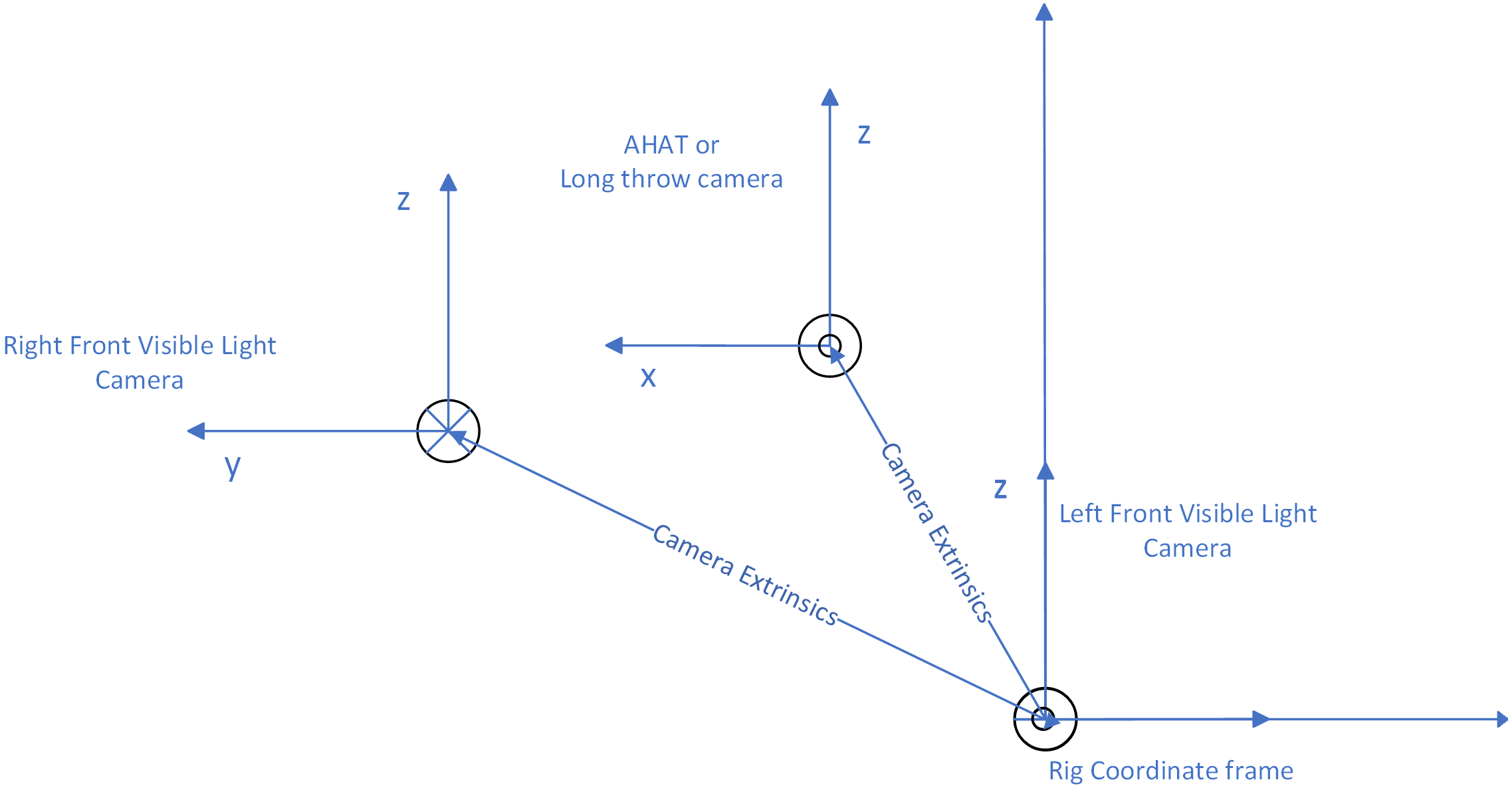}
   \end{center}
      \caption{Camera coordinate frames relative to the \emph{rigNode} (device origin -- Left Front Visible Light camera).}
   \label{fig:extrinsics}
   \end{figure}

All sensors are positioned	in a device-defined coordinate frame, which is defined as the \emph{rigNode}.
Each sensor returns its transform to the \emph{rigNode} (device origin) expressed as an extrinsics rigid body transform (rotation and translation).
On HoloLens~2, the device origin corresponds to the Left Front Visible Light Camera; therefore, the transformation returned by this sensor
corresponds to the identity transformation. Figure~\ref{fig:extrinsics} shows camera coordinate frames relative to the \emph{rigNode}.

The extrinsics transform for a camera sensor can be retrieved as follows:
\begin{lstlisting}
IResearchModeCameraSensor *pCameraSensor;
DirectX::XMFLOAT4X4 cameraPose;
// Initialize camera sensor...
// Get matrix of extrinsics wrt the rigNode
pCameraSensor->GetCameraExtrinsicsMatrix(&cameraPose);
\end{lstlisting}

In order to locate sensors with respect to other coordinate systems (\eg ``in the world"), we use the HoloLens Perception APIs~\cite{perception}.
These APIs leverage the HoloLens head tracker to locate the device in a given coordinate system (for example, defined by the location of the user at app launch time).
In order to locate the \emph{rigNode}, we need to identify it with a GUID.
This GUID is provided by the {\small\ttfamily IResearchModeSensorDevicePerception} interface:
\begin{lstlisting}
using namespace winrt::Windows::Perception::Spatial;
using namespace winrt::Windows::Perception::Spatial::Preview;

HRESULT hr = S_OK;
SpatialLocator locator;
IResearchModeSensorDevicePerception* pSensorDevicePerception;
GUID guid;
hr = m_pSensorDevice->QueryInterface(
            IID_PPV_ARGS(&pSensorDevicePerception));
if (SUCCEEDED(hr))
{
   hr = pSensorDevicePerception->GetRigNodeId(&guid);
   locator =
      SpatialGraphInteropPreview::CreateLocatorForNode(guid);
}
// further processing, define anotherCoordSystem...
auto location = locator.TryLocateAtTimestamp(timestamp,
                              anotherCoordSystem);   
\end{lstlisting}

Note that camera sensors do not directly expose intrinsics parameters.
Instead, they expose {\small\ttfamily MapImagePointToCameraUnitPlane} and {\small\ttfamily MapCameraSpaceToImagePoint} methods to convert 3D coordinates
in the camera reference frame into 2D image coordinates, and vice versa.

\section{The HoloLens2ForCV Repository}
\label{sec:examples}
The HoloLens2ForCV repository~\cite{hl2forcv} provides a few sample UWP (Universal Windows Platform) apps showing how to access and process Research Mode input streams:
for example, how to visualize streams live on device, or how to record them and correlate them in space and time with hand and eye tracking.
Section~\ref{sec:apps} provides a brief overview of the apps collected in the repository. 
Section~\ref{sec:slam} shows how, leveraging Research Mode and adding just a few lines of code, one can prototype computer vision algorithms
like TSDF Volume Integration and SLAM.

\subsection{Sample Apps}
\label{sec:apps}
In its initial version, the HoloLens2ForCV repository collects four UWP apps.
We plan to expand this initial set over time, and welcome contributions from the community.
In particular, we provide sample code to a) visualize input streams live on device and process them using OpenCV~\cite{opencv} and b) record streams to disk for offline postprocessing.

\begin{figure}[t]
   \begin{center}
      \includegraphics[width=1.\linewidth]{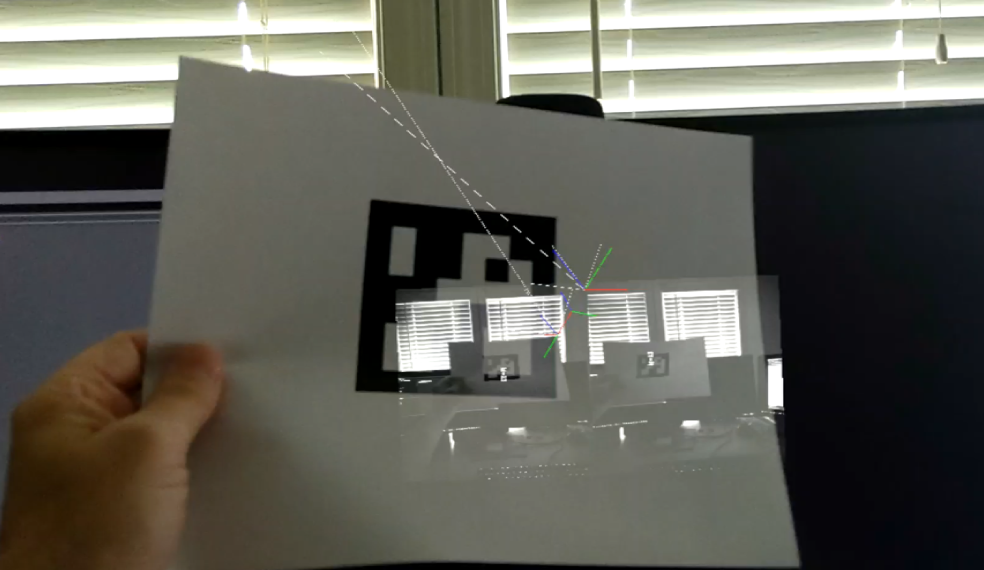}
   \end{center}
      \caption{\emph{CameraWithCVAndCalibration} app: We use OpenCV to detect arUco markers in the scene and triangulate them.}
   \label{fig:aruco}
   \end{figure}

\begin{figure}[t]
   \begin{center}
      \includegraphics[width=1.\linewidth]{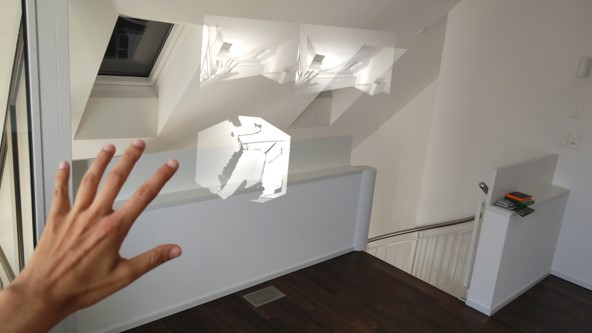}
   \end{center}
      \caption{Research Mode sensor streams visualization on device: here we show the two frontal VLC cameras and Long Throw Depth.}
   \label{fig:sensorvis}
   \end{figure}

\paragraph{Sensor Visualization and Processing:} The \emph{SensorVisualization} app allows one to visualize depth, VLC (and optionally IMU) input streams
live on device (see Fig.~\ref{fig:sensorvis}). This can be particularly useful to test, for example, sensor performance in different environments.
With the \emph{CameraWithCVAndCalibration} app, we show how to process these streams on device: we use OpenCV to detect arUco markers in the two frontal VLC cameras, and triangulate the detections (Fig.~\ref{fig:aruco}).
The \emph{CalibrationVisualization} app can be used to visualize VLC and depth coordinate frames.

\begin{figure}[t]
   \begin{center}
      \includegraphics[width=1.\linewidth]{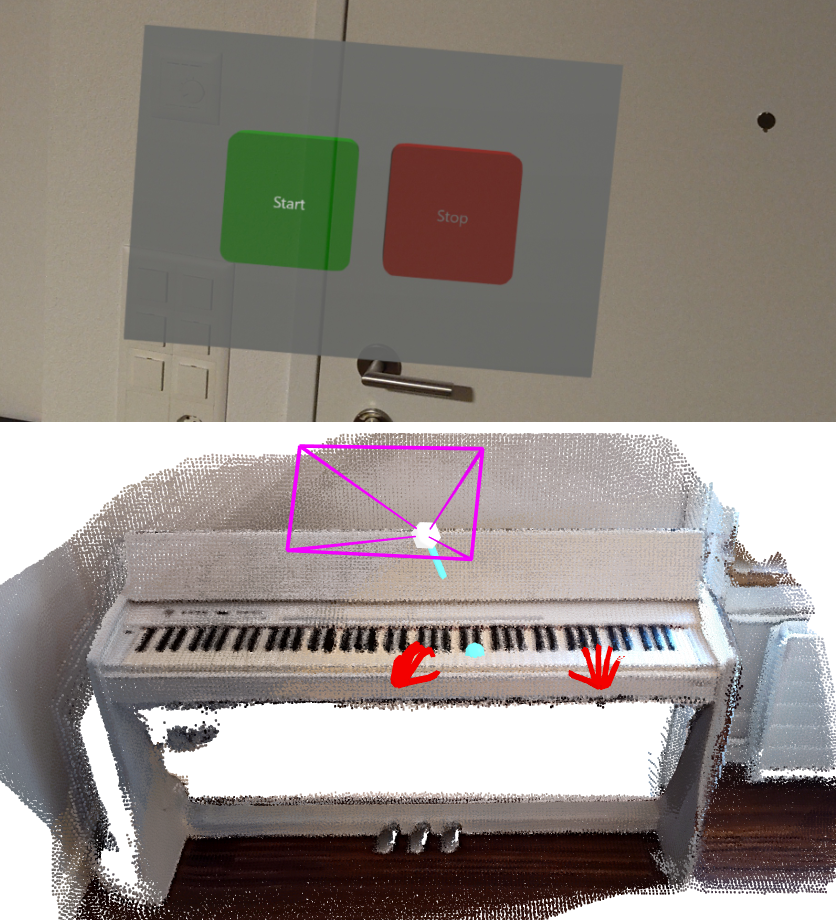}
   \end{center}
      \caption{The \emph{StreamRecorder} user interface (top). Processed outputs: Long Throw Depth frames (piano reconstruction), hand tracker output (red), eye gaze tracker output (blue sphere).}
   \label{fig:recorder}
   \end{figure}

\paragraph{Stream Recorder:} The \emph{StreamRecorder} app allows one to capture simultaneously Research Mode streams (depth and VLC),
the RGB stream from the HoloLens PV (PhotoVideo) camera, head, hand and eye tracking.
The application provides a simple user interface to start and stop capture (Fig.~\ref{fig:recorder}, top).
Streams are captured and stored on device, and can be downloaded and postprocessed by using a set of python scripts.
Namely, we provide scripts to correlate Research Mode, PV, head, hand and eye tracking streams in time and put them in a common coordinate frame.
This allows us to reconstruct the scene in 3D and analyze how the user interacts with it by looking at their eye gaze, head and hand pose (Fig.~\ref{fig:recorder}, bottom).
It is possible to add also audio capture, accessible via Windows Media APIs~\cite{media}, though this functionality is not implemented at the time of writing.

It is worth noting that the \emph{StreamRecorder} app relies on another library, \emph{Cannon} (available on~\cite{hl2forcv}). \emph{Cannon} is a collection of wrappers and utility code
for building native mixed reality apps using C++, Direct3D and Windows Perception APIs. It can be used as-is outside Research Mode for fast and easy native development.

\subsection{Computer Vision with Research Mode}
\label{sec:slam}
By leveraging Research Mode, one can use the HoloLens as a powerful tool for computer vision and robotics research.
To showcase how easily this can be done, we provide two examples: TSDF Volume Integration and SLAM.

\begin{figure}[t]
   \begin{center}
      \includegraphics[width=1.\linewidth]{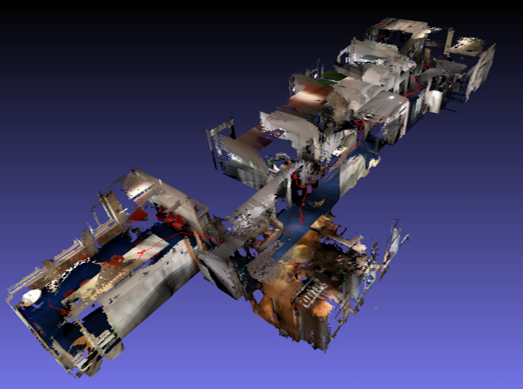}
   \end{center}
      \caption{TSDF Volume Integration~\cite{open3d} result obtained from HoloLens depth and RGB input streams.}
   \label{fig:tsdf}
   \end{figure}

\paragraph{TSDF Volume Integration:} We provide a script which runs TSDF Volume Integration using as input Long throw depth frames, RGB frames and head poses captured with the \emph{StreamRecorder} app.
We simply combine the app output with an off-the-shelf library~\cite{open3d}.
Figure~\ref{fig:tsdf} shows the reconstruction obtained by running the algorithm on almost 700 frames (head trajectory shown in red).

\paragraph{SLAM:} We feed Long Throw depth frames and RGB frames as input to a recently proposed SLAM method (BAD SLAM~\cite{badslam}). 
The camera poses estimated by the SLAM algorithm can be compared and evaluated against the head poses returned by the HoloLens.

\section{Conclusion}
\label{sec:conclusion}
We presented Research Mode for HoloLens~2, an API and a set of functionalities enabling access to the raw sensor streams on device.
Together with Research Mode, we release a public repository collecting sample apps and examples showcasing how to leverage Research Mode to build computer vision applications based on HoloLens.
With these tools, we hope to facilitate further research in the fields of computer vision and robotics, and encourage contributions from the research community.

{\small
\bibliographystyle{ieee_fullname}
\bibliography{egbib}
}

\end{document}